\documentclass[a4paper,twoside]{article}

\usepackage{epsfig}
\usepackage{subfigure}
\usepackage{calc}
\usepackage{amssymb}
\usepackage{amstext}
\usepackage{amsmath}
\usepackage{amsthm}
\usepackage{multicol}
\usepackage{pslatex}
\usepackage{apalike}
\usepackage{hyperref}
\usepackage{float}
\usepackage{balance}
\usepackage{SCITEPRESS}     

\begin{document}
\balance
\title{Autonomous Cars: Vision based Steering Wheel Angle Estimation}


\author{\authorname{Kemal Alkin Gunbay\sup{1}, Mert Arikan\sup{2} and Mehmet Turkan\sup{1}}
\affiliation{\sup{1}Department of Electrical and Electronics Engineering, Izmir University of Economics, Izmir, Turkey}
\affiliation{\sup{2}Department of Software Engineering, Izmir University of Economics, Izmir, Turkey}
\email{\{alkin.gunbay, mert.arikan\}@std.izmirekonomi.edu.tr, mehmet.turkan@izmirekonomi.edu.tr}
}

\keywords{Autonomous cars, Self-driving cars, Steering wheel angle estimation, Image processing, Deep learning.}

\abstract{Machine learning models, which are frequently used in self-driving cars, are trained by matching the captured images of the road and the measured angle of the steering wheel. The angle of the steering wheel is generally fetched from steering angle sensor, which is tightly-coupled to the physical aspects of the vehicle at hand. Therefore, a model-agnostic autonomous car-kit is very difficult to be developed and autonomous vehicles need more training data. The proposed vision based steering angle estimation system argues a new approach which basically matches the images of the road captured by an outdoor camera and the images of the steering wheel from an onboard camera, avoiding the burden of collecting model-dependent training data and the use of any other electromechanical hardware.}

\onecolumn \maketitle \normalsize \vfill

\section{\uppercase{Introduction}}
\label{sec:introduction}

\noindent Regular cameras placed outside of a vehicle are widely used for the purpose of accomplishing autonomous tasks. Captured images of the road from these outdoor cameras are processed and analyzed in machine learning and/or image processing algorithms to estimate the geometry of the road coupled with the steering parameters of the vehicle. These techniques need to capture images of the road which are precisely synchronized with the angle of the steering wheel usually gathered from a well-calibrated steering angle sensor. Moreover, the measured angle of the steering wheel should be as precise as possible in order to train successful models for autonomous tasks. The main problem here is that the collected steering angle data is highly dependent on the physical shape, and the mechanical and steering hardware of the vehicle at hand. Therefore, every vehicle model needs its own record of steering angle training data collected and specialized to the physical aspects and mechanical characteristics of that vehicle. In order to avoid the burden of collecting model-dependent training data (from scratch) per vehicle model, some mathematical tools have been emerged to tune the steering angle values between different vehicles as formulated in Equation~\eqref{eq1} such that
\begin{equation}\label{eq1}
R= \dfrac{s}{\sqrt{2-2cos(2a/n)}},  
\end{equation}
where $R$ corresponds to radius of curvature, $a$ is the steering wheel angle, $n$ is steering ration and $s$ denotes the steering wheel base. Nevertheless, some physical aspects of the vehicle are necessary parameters in this formulation.

This study proposes and argues a novel vision based structure to estimate the steering wheel angle of a vehicle regardless of its physical aspects and mechanical hardware characteristics. In this way, it is possible to obtain a global machine learning based steering wheel angle estimation model that is highly applicable for different types of vehicles without any need of collecting more data. Another important point here is that the high costs of electromechanical steering angle sensors. Although these sensors are needed only for training purposes, they increase the market prices of such self-driving vehicles. The proposed vision based system makes the estimation of the steering angle cheaper while keeping its accuracy as precise as possible.

In summary, this paper describes a novel framework which couples an outdoor camera and an onboard camera as two individual, but precisely synchronized image capturing subsystems. While the outdoor camera captures the images of the road continuously, the onboard camera records the behavior of the steering wheel which is mechanically projected to another plane that can be planar or circular. Although more efficient extensions or variations to this proposal are possible, this kind of projection helps avoid interruptions from the driver during driving, hence provides the continuum of the steering wheel capturing subsystem. In the machine learning part, this study offers training the state-of-the-art machine learning models, namely convolutional neural networks (CNNs), for the purpose of improving the efficiency of self-driving vehicles by means of these aforementioned image capturing subsystems, rather than using of any other electromechanical hardware. It is worth noting here that, CNNs are widely used in industry for very broad range of practical modeling purposes. Hence, the proposed system in this paper is also based on CNNs, although this framework needs collecting enormous numbers of training images of the road and synchronized data for the steering angle information.

This paper is organized as follows. Section 2 first reviews the literature and discusses the similarities and differences between currently existing studies and the proposed framework. Then, Section 3 explains the proposed system and details its working principles. Finally, Section 4 concludes the paper.

\section{\uppercase{Literature Review}}

\noindent There are implementations in literature in which the captured images of the steering wheel of a vehicle are processed via image processing techniques for different purposes. These implementations generally use common image processing algorithms instead of machine learning models which are indeed more convenient for real-time purposes. Basically, regular image processing techniques, which use predefined filters, are slower than machine learning models using those filters at an optimum level. In the last decade, deep neural networks (deep learning) as improved versions of artificial neural networks have become very popular in academic and industrial applications of machine learning for object classification and recognition, (biomedical) image and video analysis, speech and natural language processing, and face detection and recognition. A specific and indeed very successful application of deep neural networks, namely CNNs are widely used and mostly applied to analyze visual contents. CNNs automatically extract hierarchical attributes (features) from visual datasets in large volumes with parallel calculation using GPUs, and there is huge amount of research going on for possible practical applications of CNNs together with serious industry investments.

For the purpose of successful self-driving cars, many researchers focused on estimating the steering angle from the road images by using one or more cameras. There is a successful example of machine learning application which uses CNNs to estimate the geometry of the road accomplished by a research group for DARPA's self driving car project in~\cite{(Bojarski2016}. Unlike all other studies, they are actually estimating curvature of the road and converting it into steering angle form. Even though, this method seems like a solution to the problem of physical shape dependency, they still have to convert curvature of the road to the steering angle which requires dimensions of the car as variables.

Similarly, the system in~\cite{Meganathon2018} estimates the steering angle for self-driving cars by using processed images of the road region captured by a camera. The proposed system is close to the idea of this paper in terms of estimating steering angle but their system is based on basic computer vision techniques, which is not optimized as in CNNs. Here, the main problem still persists as their proposed system of estimating the steering angle depends on physical shape of the car, hence this is yet another model-dependent application.

Some of other studies in literature are also aimed at capturing the steering angle for a number of purposes. One of the similar applications to the proposed system in this paper is the training of a steering wheel for medical purposes described in~\cite{Lee2012}. Their system consists of several subsystems including an image capturing subsystem. A monochrome camera captures images of the steering wheel and sends them to a PC. Their purpose is to develop a measurement system to easily analyze attributes of movement pattern for a steering exercise, which is a training session for patients of any brain injury. Hence, this study only aims at comparing the position of the steering wheel to other acquired data.

There are also papers and developed methods those use an onboard camera in the vehicle for many variety of reasons. As an example study in~\cite{CoetzerandHancke}, a camera that is mounted inside the car records the face of the driver for monitoring the fatigue level through detecting and analyzing the eyes of the driver. In a similar work in~\cite{RAhmed2014}, a system is described using an onboard camera to detect the face of the driver to determine whether the driver is focused or not, by applying image processing techniques. Strangely, this system do not utilize any neural networks model. An earlier work in~\cite{Ribaric2010} indeed develops a similar system to detect the fatigue level of the driver by using CNNs to successfully estimate angle of the head rotation of the driver. Another study described in~\cite{Pyo2017} illustrates that with CNNs, it is possible to measure angle of an object very precisely by using a camera. According to their reported results, they obtain an estimation accuracy of the rotation angle of an object less than $0.154$ degrees. Additionally in~\cite{Lu2010}, authors successfully measure distance and inclination angle of an object using a camera.

In addition to the camera based steering angle measurement or estimation systems, there are a wide range of sensor based techniques. One of the modern systems that is based on Giant Magnetic Resistance (GMR) sensors~\cite{Zacharia2017} claims that one can measure the steering wheel angle reliably. There are a few more applications with GMR as in~\cite{Xiao2016}. Both of these sensor based methods can measure the steering wheel angle with high precision only when these sensors are well-calibrated and the measured angle values still depend on the mechanical parts and characteristics of the vehicle.

To summarize here, to the best of our knowledge, there is no any similar application to the proposed idea of this position paper. There are indeed systems which suggest using an onboard camera for different purposes, but not for the purpose of estimating the steering wheel angle. Moreover, this framework offers a novel idea of coupling an outdoor camera (road images) with the onboard camera (steering wheel related images) in order to develop a vehicle-independent CNNs model for extending the capability of autonomous cars. Note also that there are successful applications, as mentioned above, demonstrating that the goal of the proposed framework is highly possible to achieve, and the result will be very useful for both driver assistance and self-driving systems.

\section{\uppercase{Methodology}}

\noindent The main aim is to estimate the steering wheel angle by means of an onboard camera instead of using a steering angle sensor. To accomplish this task, images of the steering wheel needs to be captured by the onboard camera which is precisely synchronized with an outdoor camera that is responsible of capturing road images. These two image capturing subsystems need to be reliable, accurate, precise and continuous while collecting the training data.

\begin{figure}[!t]
	\vspace{-0.2cm}
	\centering
	{\epsfig{file = 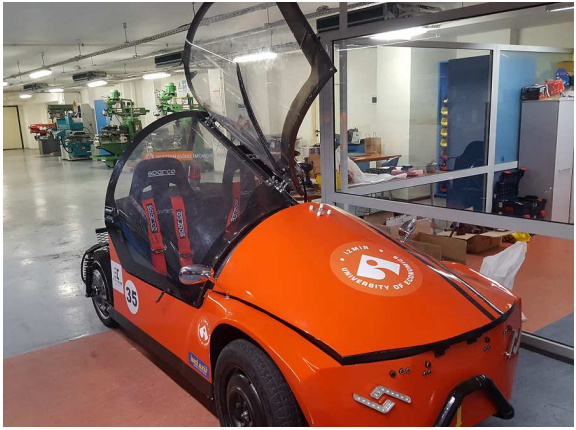, width = \linewidth}}
	\caption{The electrical car designed and manufactured in our university that will serve a basis for a testing framework of the proposed system.}
	\label{fig:car}
	\vspace{-0.1cm}
\end{figure}

To build a prototype system, the electrical car designed and manufactured in our university as illustrated in Figure~\ref{fig:car} is taken as a basis for the development and testing framework. The remaining part of this section explains the proposed system in terms of its mechanical components, as well as image capturing and machine learning parts, and details the current development state and working principles.

\subsection{Mechanical Components}

\begin{figure}[!t]
	\vspace{-0.2cm}
	\centering
	{\epsfig{file = 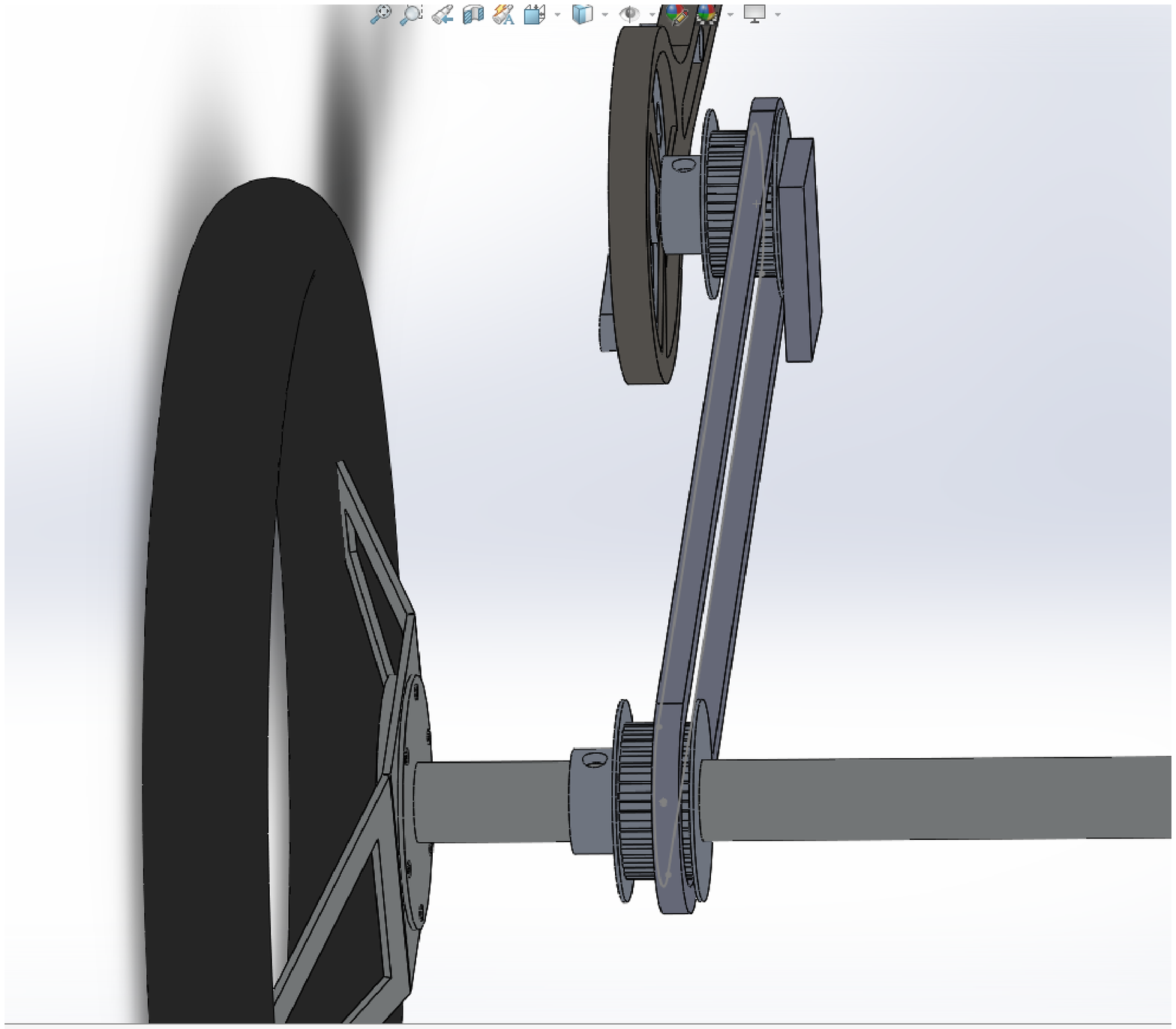, width = 5.5cm}}
	{\epsfig{file = 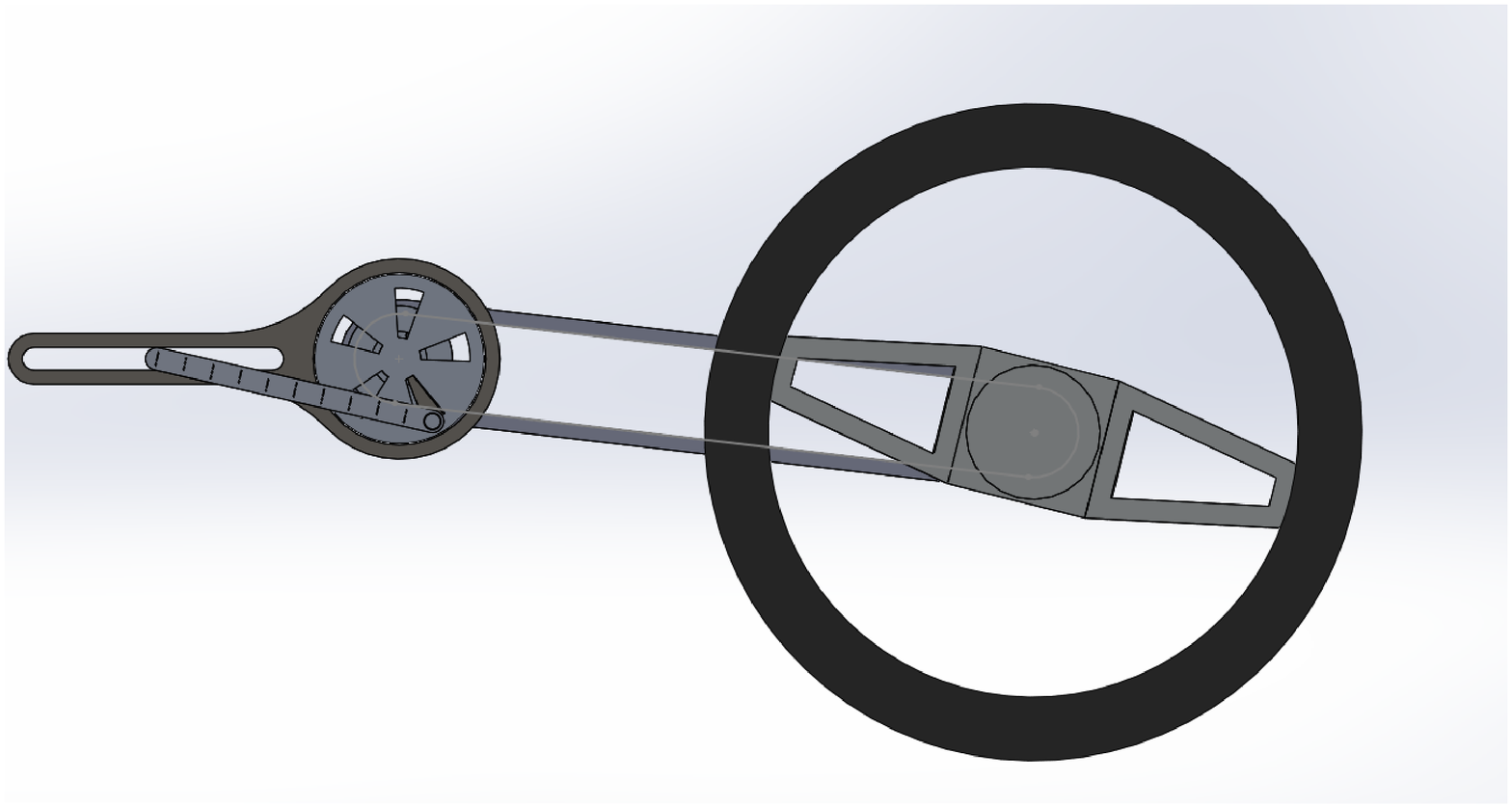, width = 5.5cm}}
	\caption{Mechanical drawings of the rotary to linear motion converter.}
	\label{fig:mech_draw}
	\vspace{-0.1cm}
\end{figure}

\begin{figure}[!t]
	\vspace{-0.2cm}
	\centering
	{\epsfig{file = 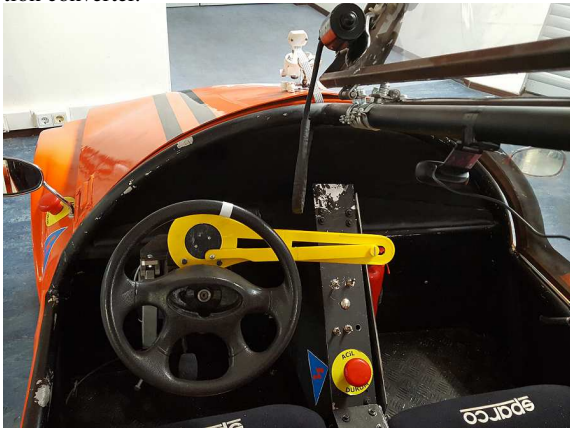, width = \linewidth}}
	\caption{A prototype motion converter is connected to the steering wheel.}
	\label{fig:converter}
	\vspace{-0.1cm}
\end{figure}

It is very important to provide the continuity of image capturing subsystems without any interruptions. The outdoor camera system needs to capture the images of the road, and the onboard camera system needs to record the behavior of the steering wheel. While more efficient variations to this current state are possible, the behavior of the steering wheel is mechanically projected to another plane that is linear. Nevertheless, this kind of projection helps avoid interruptions from the driver during driving, hence provides the continuum of the steering wheel capturing subsystem. So, the rotary motion of the steering wheel is converted into a linear motion by means a simple mechanical part as demonstrated in Figure~\ref{fig:mech_draw}. In these mechanical drawings, a timing belt and two pulleys are illustrated. One of the pulleys is co-central with the steering wheel, and the other one allows to build a co-central and freely movable motion converter. This converter projects $360$ degrees to transfer the full motion of the steering wheel, through the lever arm of the converter which periodically traces in horizontal axes. A prototype motion converter has been printed for this purpose and connected to the steering wheel, see Figure~\ref{fig:converter}. Note also that there is a marker located on the edge of the lever arm of the converter.

\subsection{Image Capturing Subsystems}

\begin{figure}[!b]
	\vspace{-0.2cm}
	\centering
	{\epsfig{file = 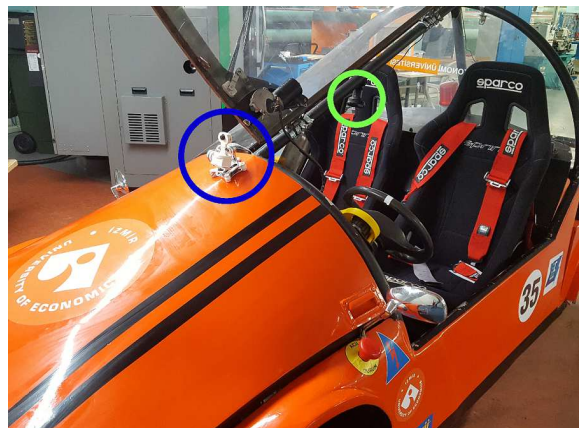, width = \linewidth}}
	\caption{A prototype of the locations of outdoor and onboard cameras.}
	\label{fig:cameras}
	\vspace{-0.1cm}
\end{figure}

In the current prototype, outdoor camera with a gimbal is located on the engine bonnet and the onboard camera is mounted inside where the field of view of this camera entirely covers the linear motion of the converter as shown in Figure~\ref{fig:cameras}. While capturing the images of the road from the outdoor camera, the motion of converter is recorded by the onboard camera. To collect synchronized training data from these two subsystems (for a CNNs model), some related meta-data should also be collected in addition to the captured images. These meta-data includes the temperature and voltage/current levels of batteries, speed of the car, and so on. Most crucially, steering angle measurements of a well-calibrated angle sensor is needed to verify the estimated steering angles of the proposed system. Custom image class and appropriate mechanism to save objects of this class are implemented in order to keep images and related meta-data intact.

\subsection{Machine Learning and Experiments}

With the collected training images from two synchronized subsystems and specialized meta-data, a CNNs model will be constructed and trained similar to~\cite{(Bojarski2016}. The proposed system will output a direct value of the steering angle, and it does not require any recalibration once it is trained successfully. More importantly, this CNNs model can possibly be used in a different car since the model parameters are not dependent on the physical aspects and hardware characteristics of the training vehicle. Moreover, possible straightforward extensions include adopting transfer learning applications for different types of vehicles such as trucks and buses.

\begin{figure}[!b]
	\vspace{-0.2cm}
	\centering
	{\epsfig{file = 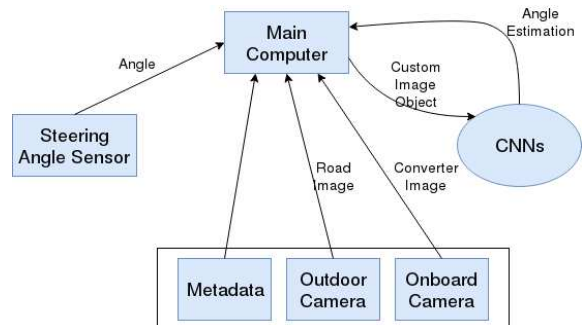, width = \linewidth}}
	\caption{Data flow diagram of the training and validation system.}
	\label{fig:Train_system}
	\vspace{-0.1cm}
\end{figure}

\begin{figure}[!t]
	\vspace{-0.2cm}
	\centering
	{\epsfig{file = 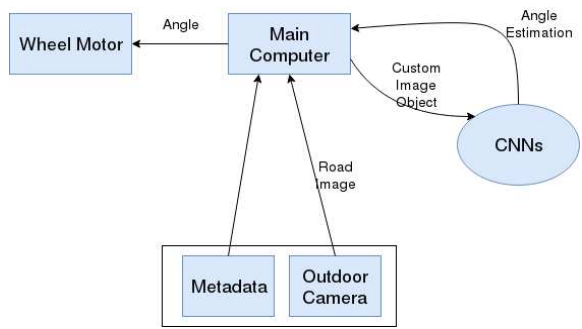, width = \linewidth}}
	\caption{Data flow diagram of the practical system.}
	\label{fig:system}
	\vspace{-0.1cm}
\end{figure}

General data flow diagram of the proposed training and validation system is given in Figure~\ref{fig:Train_system}. The onboard main computer is NVIDIA Jetson TX2. In the training phase, the captured images from two cameras and associated meta-data are sent to this main computer and saved frame-by-frame as a custom image object. Later, the training can be accomplished offline using a more powerful computer than the onboard computer. The trained CNNs model is then validated by comparing CNNs angle estimations with the measurements of a highly accurate steering angle sensor (Bosch). This is only for one time verification of the proposed system and the CNNs model, hence later on there will not be any need of such a sensor. Additionally, the data flow diagram of the practical system is illustrated in Figure~\ref{fig:system}. In this scheme there is no onboard camera since the tests are carried out analyzing the images captured by the outdoor camera only, and the associated meta-data if necessary. The estimated steering angle value is then directed to the wheel motor per frame.

\section{\uppercase{Conclusion}}
\label{sec:conclusion}

\noindent Independence from the physical aspects of the vehicle is crucial for modular, model-agnostic autonomous car systems and machine learning algorithms applied in those systems. Instead of using an expensive and hardware dependent steering angle sensor, the proposed vision based system offers a novel and cost effective solution to the problem of estimating the steering wheel angle. The proposed structure further avoids the burden of collecting model-dependent training data. Development and implementation of this system is currently an ongoing work and its statistical accuracy needs to be analyzed through real-time tests.

\vfill
\bibliographystyle{apalike}
{\small
\bibliography{example}}

\begin{thebibliography}{}

\bibitem[Ahmed et~al., 2014]{RAhmed2014}
Ahmed, R., Emon, K. E.~K., and Hossain, M.~F. (2014).
\newblock Robust driver fatigue recognition using image processing.
\newblock In {\em Int. Conf. Informatics, Electronics Vis.}, pages 1--6.

\bibitem[Bojarski et~al., 2016]{(Bojarski2016}
Bojarski, M., Testa, D., Dworakowski, D., Firner, B., Flepp, B., Goyal, P.,
  Jackel, L., Monfort, M., Muller, U., Zhang, J., Zhang, X., Zhao, J., and
  Zieba, K. (2016).
\newblock End to end learning for self-driving cars.
\newblock {\em CoRR, 1604.07316}.

\bibitem[Coetzer and Hancke, 2011]{CoetzerandHancke}
Coetzer, R.~C. and Hancke, G.~P. (2011).
\newblock Eye detection for a real-time vehicle driver fatigue monitoring
  system.
\newblock In {\em IEEE Intell. Vehicles Symp. (IV)}, pages 66--71.

\bibitem[Lee et~al., 2012]{Lee2012}
Lee, H.~M., Li, P.~C., Wu, S.~K., and You, J.~Y. (2012).
\newblock Analysis of continuous steering movement using a motor-based
  quantification system.
\newblock {\em Sensors}, 12(12):16008--16023.

\bibitem[Lu et~al., 2010]{Lu2010}
Lu, M., Hsu, C., and Lu, Y. (2010).
\newblock Distance and angle measurement of distant objects on an oblique plane
  based on pixel variation of {CCD} image.
\newblock In {\em IEEE Instr. Measurement Tech. Conf. Proc.}, pages 318--322.

\bibitem[Meganathan et~al., 2018]{Meganathon2018}
Meganathan, R.~R., Kasi, A.~A., and Jagannath, S. (2018).
\newblock Computer vision based novel steering angle calculation for autonomous
  vehicles.
\newblock In {\em IEEE Int. Conf. Robotic Comput.}, pages 143--146.

\bibitem[Pyo and Kim, 2017]{Pyo2017}
Pyo, J. and Kim, K. (2017).
\newblock Precise angle estimation using geometry features for bin picking.
\newblock In {\em Int. Conf. Adv. Robotics}, pages 281--283.

\bibitem[Ribaric et~al., 2010]{Ribaric2010}
Ribaric, S., Lovrencic, J., and Pavesic, N. (2010).
\newblock A neural-network-based system for monitoring driver fatigue.
\newblock In {\em IEEE Mediter. Electrotech. Conf.}, pages 1356--1361.

\bibitem[Xiao, 2016]{Xiao2016}
Xiao, M. (2016).
\newblock A high-accuracy steering wheel angle sensor based on {GMR}.
\newblock In {\em Int. Conf. Instr. Measurement, Computer, Commun. Control},
  pages 1--4.

\bibitem[Zacharia et~al., 2017]{Zacharia2017}
Zacharia, S., George, T., Rufus, E., and Alex, Z.~C. (2017).
\newblock Implementation of steering wheel angle sensor system with controlled
  area network.
\newblock In {\em Int. Conf. Intell. Comput. Instr. Control Tech.}, pages
  54--60.

\end{thebibliography}

\vfill
\end{document}